\documentclass{ecai} 
\usepackage{graphicx}
\usepackage{latexsym}
\usepackage{amssymb}
\usepackage{amsmath}
\usepackage{amsthm}
\usepackage{booktabs}
\usepackage{enumitem}
\usepackage{graphicx}
\usepackage{color}
\usepackage{epsfig} 
\newcommand{\BibTeX}{B\kern-.05em{\sc i\kern-.025em b}\kern-.08em\TeX}

\begin{document}

%%%%%%%%%%%%%%%%%%%%%%%%%%%%%%%%%%%%%%%%%%%%%%%%%%%%%%%%%%%%%%%%%%%%%%%%

\begin{frontmatter}

%%% Use this command to specify your submission number.
%%% In doubleblind mode, it will be printed on the first page.

\paperid{404} 

%%% Use this command to specify the title of your paper.

\title{SPIN: SE(3)-Invariant Physics Informed Network for Binding Affinity Prediction}

%%% Use this combinations of commands to specify all authors of your 
%%% paper. Use \fnms{} and \snm{} to indicate everyone's first names 
%%% and surname. This will help the publisher with indexing the 
%%% proceedings. Please use a reasonable approximation in case your 
%%% name does not neatly split into "first names" and "surname".
%%% Specifying your ORCID digital identifier is optional. 
%%% Use the \thanks{} command to indicate one or more corresponding 
%%% authors and their email address(es). If so desired, you can specify
%%% author contributions using the \footnote{} command.

%\author[A]{\fnms{Anonymous}~\snm{Author}}
% \author{
%     Anonymous Author
% }
\author[A]{\fnms{Seungyeon}~\snm{Choi}\footnote{First Author. Email: tmddus1553@yonsei.ac.kr}}
\author[A]{\fnms{Samgmin}~\snm{Seo}}
\author[A]{\fnms{Sanghyun}~\snm{Park}\thanks{Corresponding Author. Email: sanghyun@yonsei.ac.kr}} 

\address[A]{Department of Computer Science, Yonsei University, Seoul, Republic of Korea}
%%% Use this environment to include an abstract of your paper.

\begin{abstract}
Accurate prediction of protein-ligand binding affinity is crucial for rapid and efficient drug development. Recently, the importance of predicting binding affinity has led to increased attention on research that models the three-dimensional structure of protein-ligand complexes using graph neural networks to predict binding affinity. However, traditional methods often fail to accurately model the complex’s spatial information or rely solely on geometric features, neglecting the principles of protein-ligand binding. This can lead to overfitting, resulting in models that perform poorly on independent datasets and ultimately reducing their usefulness in real drug development. To address this issue, we propose \textbf{SPIN}, a model designed to achieve superior generalization by incorporating various inductive biases applicable to this task, beyond merely training on empirical data from datasets. For prediction, we defined two types of inductive biases: a geometric perspective that maintains consistent binding affinity predictions regardless of the complex’s rotations and translations, and a physicochemical perspective that necessitates minimal binding free energy along their reaction coordinate for effective protein-ligand binding. These prior knowledge inputs enable the \textbf{SPIN} to outperform comparative models in benchmark sets such as CASF-2016 and CSAR HiQ.
Furthermore, we demonstrated the practicality of our model through virtual screening experiments and validated the reliability and potential of our proposed model based on experiments assessing its interpretability.
\end{abstract}

\end{frontmatter}

%%%%%%%%%%%%%%%%%%%%%%%%%%%%%%%%%%%%%%%%%%%%%%%%%%%%%%%%%%%%%%%%%%%%%%%%

\section{Introduction}

The binding affinity (BA) of a small molecule to a target protein associated with a disease indicates whether the small molecule (a ligand) can become a new drug \cite{hopkins2014role,parenti2012advances}. The higher the BA, the stronger the binding to the target protein and the desired effect, which can be measured using biological tests. However, these methods are expensive and  time consuming. 
% Therefore, the prediction of BA using machine learning (ML) and deep learning (DL) holds significant potential for accelerating the drug development process and reducing costs, compared to the critical limitations associated with biological testing for measuring BA \cite{wang2024prediction}. 
Therefore, compared with biological testing, BA prediction using machine learning (ML) and deep learning (DL) may accelerate drug development and reduce costs \cite{wang2024prediction}.
Powered by this need, numerous studies have been conducted to predict BA, especially with a recent increase in efforts to predict BA by analyzing the three-dimensional structure of complexes formed between proteins and ligands \cite{meli2022scoring}. 
This trend stems from research indicating that since the BA between a protein and a ligand is based on their interaction in 3D space, predicting BA through information derived from the three-dimensional structural characteristics of the protein and ligand is a rational approach \cite{batool2019structure,lionta2014structure}.

% Methods for representing the 3D structure of a Protein-ligand complex for the purpose of predicting binding affinity can be broadly classified into two categories: representing the complex in a grid form and representing it in a graph form. First, the grid representation method involves voxelizing the structure of the protein-ligand complex into a 3D grid and then typically using 3D CNNs to predict the binding affinity. However, most grid forms produced through voxelization do not convey the structural information of the complex, making CNN architectures relatively inefficient in describing the chemical structure of the protein-ligand complex. Additionally, using a 3D rectangular grid representation can create high-dimensional sparse 3D matrices, which may involve high computational costs. To address these issues, recent studies have emerged that predict binding affinity by representing the complex structure in graph form. This involves defining the cartesian coordinates of each atom composing the complex as position information and stating that two different atoms have an edge if the distance between them is within a certain value. Previous research has sought to model the geometric information of the protein-ligand complex in three-dimensional space through such representation methods. In particular, PigNet was able to enhance the model's generalization performance by defining a set of physical laws that encompass the geometric information of the protein-ligand complex as an inductive bias for the prediction model. 

DL models utilize various methods to represent the 3D structures of protein–ligand complexes for BA prediction and are primarily categorized into grid- and graph-based representations.
The grid-based approach involves voxelizing the structure of the protein-ligand complex into a 3D grid format, after which it is commonly processed using 3D convolutional neural network (CNN) to predict the binding affinity \cite{stepniewska2018development,zheng2019onionnet}.
\begin{figure}
	\centering
    \includegraphics[width=3.37in]{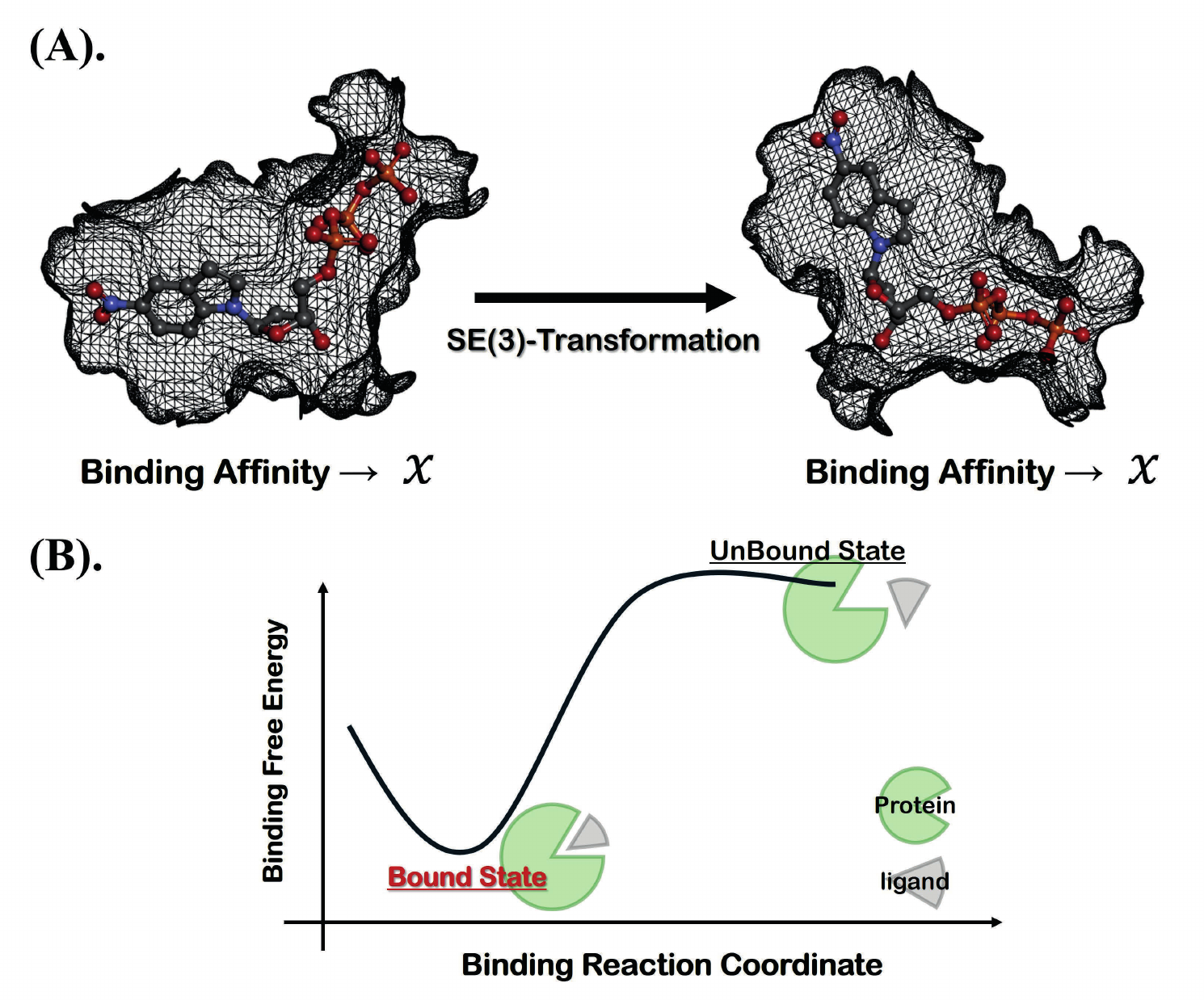}
%    \includegraphics[width=3.37in]{figs/Intro_figure_incline.jpg}
	% \caption{The two inductive biases of the protein-ligand binding affinity prediction model. (A). The geometric inductive bias that the binding affinity of a complex remains the same even after undergoing SE(3)-transformations such as translations and rotations (B). The physicochemical inductive bias that the protein-ligand complex to be predicted for binding affinity should have its binding free energy minimized along their reaction coordinate.}
    \caption{Two inductive biases of the protein–ligand BA prediction model: (A). Geometric inductive bias, i.e., the BA of a complex remains constant despite undergoing SE(3)-transformations, such as translations and rotations. (B). Physicochemical inductive bias, i.e., protein-ligand complex should be positioned at the point where binding free energy is minimal among their possible reaction coordinates.}
	\label{fig1}
\end{figure}
However, most grid-based representations do not completely encapsulate the structural information of the complexes. Consequently, the CNN architecture becomes relatively inefficient in depicting the geometric intricacies of protein-ligand complexes. 
Furthermore, employing a 3D rectangular grid representation can generate high-dimensional sparse 3D matrices, potentially resulting in significant computational costs.
To address these issues, recent research has focused on representing the structure of complexes in a graphical format to predict BA. This approach defines the atoms of the protein and ligand as nodes in a graph and assigns edge information to atom pairs that are within a certain distance threshold. Consequently, the complex is represented as either a single or multiple graphs, enabling the operation of graph neural network models \cite{li2021structure,nguyen2021graphdta,song2020communicative,wu2021representing}. However, traditional methods that consider only the connectivity information between atoms fail to adequately model the spatial information of complexes. Additionally, the need to define numerous preset rules to handle complex geometric features limits the model's flexibility when encountering unseen samples during training.
Therefore, designing a model that efficiently models the complex information of the protein-ligand complex, ensuring it has excellent generalization capabilities, is essential.

Applying the concept of Physics-Informed Neural Network (PINN) \cite{hao2022physics,karniadakis2021physics} to BA prediction can be a promising approach to enhancing model generalization performance by incorporating domain knowledge inherent in the protein-ligand complex into the prediction model. 
Specifically, beyond relying solely on the empirical distribution of data for model training, defining the immutable governing laws of the real world as the inductive bias of the prediction model ensures robustness against noise in the training dataset. Furthermore, the same domain knowledge can be applied to unseen samples during training to achieve superior generalization performance \cite{cui2023knowledge}. 
In this research, the governing laws, or inductive biases, are defined in two categories as illustrated in figure 1. The first inductive bias, as shown in figure 1.(A), is based on the fundamental principle that the binding affinity of a protein-ligand complex remains constant despite SE(3)-transformations, such as rotations and translations. This inherently sensible law allows for more efficient modeling of the spatial information of the complex compared to some binding affinity prediction models that only input three-dimensional coordinates. The second inductive bias, illustrated in figure 1.(B), relies on the physicochemical principle that the binding free energy of a protein-ligand complex is located at the point where the energy is minimized among all possible geometric configurations of the protein and ligand \cite{woo2005calculation}. These two governing laws represent domain knowledge that can be explained within the context of protein-ligand complexes and are immutable truths applicable equally in both training and testing scenarios.
In this way, we propose \textbf{SPIN}, a graph transformer model that incorporates the SE(3)-invariant principle, which is invariant to rotation and translation, and the principle of minimal binding free energy, as its geometric and physicochemical inductive biases, respectively, to achieve high generalization performance.
% To the best of our knowledge, this approach of defining and infusing two real-world inductive biases into a binding affinity prediction model represents the first attempt of its kind within this task.
Consequently, we have validated that the \textbf{SPIN} outperforms other prediction models on two benchmark datasets. Extensive experiments have also allowed us to assess the interpretability and practicality of our proposed model.

\noindent{Our main contributions can be summarized as follows:}
% \begin{itemize}
% \item We propose \textbf{SPIN}, a novel binding affinity prediction model that incorporates various prior knowledge to achieve superior generalization performance.

% \item To the best of our knowledge, this is the first model for predicting binding affinity where geometric knowledge and physicochemical knowledge are defined as inductive biases.

% \item \textbf{SPIN} achieves state-of-the-art (SOTA) performance in the CASF-2016 and CSAR-HiQ benchmarks, demonstrating the model's excellent interpretability and practicality.
% \end{itemize}
\begin{itemize}
\item We propose \textbf{SPIN}, a novel binding affinity prediction model that is infused with essential prior knowledge applicable to various protein-ligand interactions, designed to achieve superior generalization performance.

\item To the best of our knowledge, this is the first model for predicting binding affinity where geometric and physicochemical knowledge are defined as inductive biases.
This approach integrates empirical learning from deep learning with immutable truths applicable in reality, allowing for extreme efficiency with limited data. Furthermore, The trained prediction model inherently reflects the fundamental principles of protein-ligand binding. 

\item \textbf{SPIN} achieves state-of-the-art (SOTA) performance in the CASF-2016 and CSAR-HiQ benchmarks, demonstrating the model's excellent interpretability and practicality through various experiments.
\end{itemize}

\newpage
\section{Related Work}
\subsection{Structure-based BA Prediction}
The prediction of protein–ligand BA is indispensable in drug screening, wherein the selection of molecular structures that are viable as drug candidates from numerous molecular structures is crucial \cite{parenti2012advances}. Recognizing this importance, numerous studies have proposed various methodologies for BA prediction owing to the advancements in ML and DL. Moreover, recent research, such as that involving AlphaFold \cite{jumper2021highly,varadi2023impact}, which models the tertiary structure of proteins, has significantly enhanced the potential and possibility of studies that aim to predict binding affinities by modeling the three-dimensional structures of protein-ligand complexes \cite{wang2024structure}. This approach can be categorized into the following three types: ML-, CNN-, and GNN-based methods. 

ML-based methods typically utilize predefined rules to extract descriptors from interactions within a certain distance between protein and ligand atoms, using ML models such as support vector regression and random forest, to predict BA \cite{ballester2010machine}. However, these methods only consider atom pair interactions between proteins and ligands, neglecting the 
 interactions within themselves and between multiple atom types. Thus, they fail to capture the rich spatial correlations in protein-ligand complexes. 
CNN-based methods transform the protein-ligand complex into a 3-D rectangular grid representation and implement a 3-D CNN model to extract features for BA prediction \cite{stepniewska2018development,zheng2019onionnet}. 
Although this grid representation enables the mining of spatial and local correlations, it generate empty grid points where no atoms exist, causing inefficient computation and unnecessary memory usage. 
Furthermore, grid representations lack distance awareness and rotational invariance, which can render the prediction performance unstable.
To address these issues, GNN-based methods have been developed for BA prediction \cite{nguyen2021graphdta,wu2021representing,lim2019predicting,moon2022pignet}. 
The GNN approaches define the atoms of proteins and ligands as nodes in a graph, and atom pairs within a certain distance threshold are connected with edge information, representing the complex as a single graph or multiple graphs for processing by GNN models. 
Recent efforts also directly input the 3D coordinates of atoms to predict binding affinity \cite{danel2020spatial}, and additional research aims to map various spatial features of the complex to more deeply learn its geometric characteristics \cite{li2021structure,li2023giant}. 
However, these methods can be sensitive to rotations and translations of the complex or may struggle to flexibly handle various geometric combinations not encountered during training due to too many predefined rules.
To overcome the absence of generalization, our goal is to introduce various inductive biases into the predictive model. These inductive biases are intended to enable the model to account for unseen geometrical configurations and ensure robustness against variations in complex orientation and position, thus enhancing the model’s predictive accuracy and generalization capabilities across different protein-ligand complexes.

\subsection{PINN}
Deep learning models have achieved tremendous success in various domains such as computer vision and natural language processing. However, the traditional approach of learning from large amounts of data still faces challenges in extracting interpretable information. Additionally, while existing data-driven approaches may align well with observations in training data, predictions from models can be physically inconsistent or result in impossible values, leading to reduced generalization performance. This issue is especially pronounced in domains like physics and fluid dynamics, where data collection is relatively difficult, unlike in fields with vast amounts of data such as image processing or natural language processing.

To address these issues, some domains are actively researching physics-informed neural network (PINN), which integrate governing physical laws and domain knowledge into deep learning models \cite{karniadakis2021physics,meng2022physics,sharma2023review,cuomo2022scientific}. Integrating domain knowledge into DL models means defining laws applicable to a domain as inductive biases during the model's training process. This ensures that the DL model implicitly satisfies these predefined laws during learning and inference. Even if the training dataset contains outliers or noise, integrating these immutable truths applicable to the real world can enhance the model's robustness. Consequently, the trained model can make more realistic predictions by training on not just empirical information from the dataset but also incorporating prior domain knowledge. This not only improves model generalization but also uses prior knowledge to illuminate the inner mechanisms of deep learning and provide theoretical insights \cite{spigler2019jamming,geiger2020scaling}. Especially in studies predicting the interactions between proteins and ligands, research has been conducted that integrates physicochemical prior knowledge related to binding free energy to apply the PINN framework \cite{moon2022pignet}. This research achieved high generalization performance in the model by defining the physical law that the positions of the atoms in the ligand within the experimentally validated protein-ligand complex structure are located at the local minimum of the potential energy as prior knowledge, i.e., inductive bias, of the prediction model. However, this model has a clear limitation in that it considers only the connection information between the protein and the ligand, failing to model the geometric information between the two structures.

\begin{figure*}
	\centering
    \includegraphics[width=\textwidth]{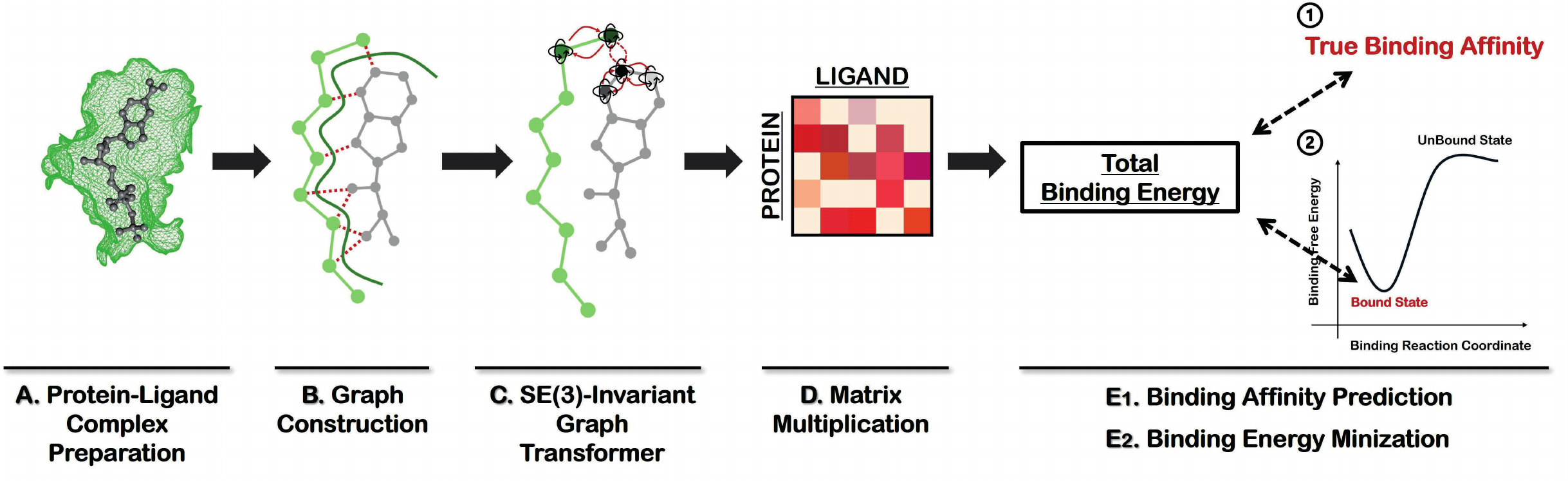}
	\caption{Overview of \textbf{\underline{SPIN}}. \textbf{A.} Protein-ligand complex preparation from PDBbind Dataset  \textbf{B.} Atoms constituting a complex are defined as nodes, and the connections between atoms are defined as edges, representing the entire structure as a single graph \textbf{C.} Each node feature is updated through an SE(3)-Graph transformer, in conjunction with the features of the connected edges, to model the geometric information in the three-dimensional space of the complex  \textbf{D.} The protein-ligand interaction matrix is extracted through matrix multiplication of the final node representation vectors of protein and ligand atoms. The computed interaction matrix is defined in terms of pairwise energy values between protein atoms and ligand atoms.  $\textbf{E}_\textbf{1}$. The binding affinity is predicted by summing the values of the extracted pairwise interaction matrix. $\textbf{E}_\textbf{2}$. The binding free energy is minimized by enforcing that the derivatives of the pairwise distances between protein and ligand atoms in the extracted pairwise interaction matrix equal zero.}
	\label{fig2}
\end{figure*}
\section{Preliminaries}
\textit{Definition 1.} \textbf{Protien-Ligand Graph Construction}
Given a protein-ligand comlex as shown in Figure2.(B), we define the feature of atom node of protein and ligand as $V^P\in \mathbb{R}^{N \times D_F}$ and $V_L\in \mathbb{R}^{N \times D_F}$ with position matrix $\mathbf{x} \in \mathbb{R}^{(N+M) \times 3}$. The protein atom features include atom type, amino acid types and whether the atoms are backbone atoms. The ligand atom features include atom type, hybridzation, formal charge, degree and whether the atoms are aromatic atoms. The \textit{knn} graph is utilized to define edges $\mathbf{e}_{ij}$ between atoms based on the 3D coordinates of each atom. $\mathbf{e}_{ij}$ is represented as a 4-dim one-hot vector, denoting connections between protein atoms, connections between ligand atoms, protein-ligand connections, and ligand-protein connections.

\noindent{\textit{Definition 2.} \textbf{Protien-Ligand Binding Affinity}} 
Our goal is to predict the binding affinity, i.e., how strongly a protein and ligand bind, given the structure of a protein-ligand complex. To achieve this, we construct and train a regression model $\phi: (V^P, V^L, \mathbf{x}) \rightarrow y$ that predicts the binding affinity using the features of the atoms composing the protein and ligand in the given complex, $V^P$, $V^L$, and the 3D coordinates of the atoms, $\mathbf{x}$.

\section{Method}
\subsection{Overview}
 To achieve a predictive model with high generalizability, it is important to define as the model's inductive bias the prior knowledge that encompasses the real-world or the datasets being used, in addition to empirical learning based on datasets. For this purpose, we inject into the predictive model inductive biases from both geometric and physicochemical perspectives. The overall framework can be seen in Figure 2. More specifically, as shown in Figure \ref{fig2}.C, it is possible to internalize a geometric inductive bias that satisfies SE(3)-invariant characteristics by utilizing the SE(3)-graph transformer, and in Figure \ref{fig2}.$\text{E}_2$, it can satisfy the physicochemical inductive bias by enforcing the condition that the binding free energy should be minimized at the given complex position. 
 
 The statement below outlines the proposition necessary for the overall framework to satisfy the characteristic of being SE(3)-invariant.
 
\noindent{\textbf{Proposition 1.}}
\textit{Denoting the SE(3)-transformation as $T_g$, let’s $f(x)$ be an SE(3)-invariant function, i.e., $f(x)=f(T_g(x))$. If function $g(x)$ is SE(3)-invariant function, i.e., $g(x)=g(T_g(x))$, then we have composition function $g(f(x))$ is also SE(3)-invariant function, i.e., $g(f(x)) = g(f(T_g(x)))$.} 

\noindent{Here, $f(x)$ refers to the SE(3)-invariant graph transformer, and $g(x)$ denotes the SE(3)-invariant prediction function that ultimately predicts the BA. This proposition indicates the conditions necessary for the proposed framework to possess SE(3)-invariant characteristics overall when performed across multiple functions. The following sections will detail the operation of an SE(3)-invariant graph transformer (Section 4.2), how the function predicting binding affinity can satisfy physical laws (Section 4.3), and whether that function is SE(3)-invariant (Section 4.3).}

\subsection{SE(3)-Invariant Graph Transformer}
We first encode the types ($V$) of the atoms composing the given protein and ligand individually through an embedding layer, as below.
\begin{equation}
\begin{aligned}
&\mathbf{h}_P^{0}=\operatorname{Linear}(V^P) \in \mathbb{R}^{N \times D_E}
\\
&\mathbf{h}_L^{0}=\operatorname{Linear}(V^L) \in \mathbb{R}^{M \times D_E}
\\
%&\mathbf{h}^{0}=[\mathbf{h}_P^{0} || \mathbf{h}_L^{0}] 
&\mathbf{h}^0=\left[\mathbf{h}_P^0 \| \mathbf{h}_L^0\right]
\end{aligned}
\end{equation}
Here, the result of concatenating $\mathbf{h}_L^{0}$ and $\mathbf{h}_P^{0}$, denoted as $\mathbf{h}^{0}$, becomes the initial hidden representation that serves as the input to the SE(3)-Invariant graph transformer. Subsequently, to update the representation of each node, we define the SE(3)-invariant graph transformer layer, which is invariant to SE(3)-transformation, as follows.
\begin{equation}
\begin{aligned}
\mathbf{h}_i^{l+1}=\mathbf{h}_i^l+\sum_{j \in \mathcal{V}, i \neq j}f_\mathbf{h}\left(\left\|\mathbf{x}_i-\mathbf{x}_j\right\|, \mathbf{h}_i^l, \mathbf{h}_j^l, \mathbf{e}_{i j} ; \theta_\mathbf{h}\right)
\end{aligned}
\end{equation}
where $h_i^l$ is the hidden representation vector for atom $i$ at the $l$th layer and $\left\|\mathbf{x}_i-\mathbf{x}_j\right\|^2$ is the euclidean distance between two atom $i$ and $j$. The update function $f_\mathbf{h}$ computes messages for updating the node state after aggregating information from neighboring nodes using attention operations as follows.
\begin{equation}
\begin{aligned} 
% & f_\mathbf{h} = \operatorname{Softmax}\left(\frac{\mathbf{q}_i \cdot \mathbf{k}_j}{\sqrt{d_k}} \right) \cdot \textbf{v}_j \cdot \sigma \operatorname{Linear}\left(\mathbf{r}_{ij}  \right)
& f_\mathbf{h} = \operatorname{\textbf{Attn}}\left(\mathbf{q}_i,\mathbf{k}_j,\textbf{v}_j \right) \cdot \operatorname{Linear}\left(\mathbf{r}_{ij}  \right)
\\& \mathbf{q}_i = \operatorname{Linear}\left(\mathbf{h}_i^{0}  \right)
\\& \mathbf{k}_j= \operatorname{Linear}\left([\mathbf{r}_{ij}|| \mathbf{e}_{ij}||\mathbf{h}_i ||\mathbf{h}_j  ]\right)
\\& \mathbf{v}_j= \operatorname{Linear}\left([\mathbf{r}_{ij}|| \mathbf{e}_{ij}||\mathbf{h}_i ||\mathbf{h}_j  ]\right)\ 
\end{aligned}
\end{equation} 
Here, $\mathbf{q}_i$, $\mathbf{k}_j$, and $\mathbf{v}j$ represent the query, key, and value matrices for the attention operation, respectively, and $\mathbf{r}_{ij}$ is defined as the distance embedding with radial basis functions located at 20 centers between 0Å and 10Å. The final atom hidden representation $\mathbf{h}^L$ becomes the node state where spatial information of neighboring atoms is aggregated, explicitly considering the relationships between protein and ligand atoms.

\subsection{SE(3)-Invariant Physics-informed Prediction}
The obtained final representation vectors of protein and ligand atoms are used to model the pairwise interaction between the protein and ligand. For conciseness, we denote the final representation vector of the atoms constituting the ligand as $\mathbf{h}^L_M$, and the atoms constituting the protein as $\mathbf{h}^L_P$. The protein-ligand binding affinity can be converted as the sum of the atom pairwise van der Waals interaction energy between the protein and the ligand \cite{du2016insights}. To achieve this, we calculate the protein-ligand interaction matrix $\mathbf{H}$ through the matrix multiplication of $\mathbf{h}_M$ and $\mathbf{h}_P$, which are outputted by the SE(3)-invariant graph transformer, as shown in Figure 4.D and as described in Equation (4). 

Additionally, we model the van der Waals interaction by calculating the pairwise distance between protein atoms and ligand atoms, along with the interaction matrix $\mathbf{H}$. The van der Waals ineteraction energy can be formulated as in Equation (5) through the Lennard-Jones potential fomula.
\begin{equation}
\begin{aligned} 
\mathbf{H} = \mathbf{h}^L_M \cdot {\mathbf{h}_P^L}^\top
\end{aligned}
\end{equation} 

\begin{equation}
\begin{aligned} 
\mathrm{E}^{\mathtt{VDW}}=\sum_{i, j} c_{i j}\left[\left(\frac{\mathbf{u}_{ij}+\mathbf{H}_{i j}}{\|\mathbf{x}_i-\mathbf{x}_j\|}\right)^{12}-2\left(\frac{\mathbf{u}_{ij}+\mathbf{H}_{i j}}{\|\mathbf{x}_i-\mathbf{x}_j\|}\right)^{6}\right]
\end{aligned}
\end{equation} 
Here, $\mathbf{u}_{ij}$ refers to the sum of the van der Waals radii of the i-th ligand atom and the j-th protein atom. The van der Waals radii for each atom type were obtained from the parameters of X-score \cite{wang2002further}. Thus, $\text{E}^{\texttt{VDW}}$ can be calculated as a combination of parameters parameterized by the SE(3)-invariant transformer and the parameters of a physics formula utilizing the positional information of the protein-ligand complex.

An important point here is that for the overall framework of binding affinity prediction to be invariant to the rotation and translation of 3D objects, the sum of the protein-ligand pairwise interactions we model must also be invariant to translation and rotation. To this end, we prove below that the interaction we model is SE(3)-invariant.

\textit{Proof.} Let $T_g\left(\mathbf{x}\right)$ can be written explicitly as $T_g\left(\mathbf{x}\right)=\boldsymbol{R} \mathbf{x} + \mathbf{b}$, where $\boldsymbol{R} \in \mathbb{R}^{3 \times 3}$ is the rotation matrix and $\boldsymbol{b} \in \mathbb{R}^3$ is the translation vector. 
\begin{equation}
\resizebox{1.\hsize}{!}{
$\begin{aligned} 
&\sum_{i, j} \left[\left(\frac{\mathbf{u}_{ij}+\mathbf{H}_{i j}}{
\left\|\mathbf{x}_i-\mathbf{x}_j\right\|^2
}\right)^{12}-\left(\frac{\mathbf{u}_{ij}+\mathbf{H}_{i j}}{\left\|\mathbf{x}_i-\mathbf{x}_j\right\|^2}\right)^6\right] \\
&=\sum_{i, j} \left[\left(\frac{\mathbf{u}_{ij}+\mathbf{H}_{i j}}{
T_g\left(\left\|\mathbf{x}_i-\mathbf{x}_j\right\|^2\right)
}\right)^{12}-\left(\frac{\mathbf{u}_{ij}+\mathbf{H}_{i j}}{
T_g\left(\left\|\mathbf{x}_i-\mathbf{x}_j\right\|^2\right)
}\right)^6\right] \\
&=\sum_{i, j} \left[\left(\frac{\mathbf{u}_{ij}+\mathbf{H}_{i j}}{
\left\|\left(\boldsymbol{R} \mathbf{x}_i+\boldsymbol{b}\right)-\left(\boldsymbol{R} \mathbf{x}_j+\boldsymbol{b}\right)\right\|^2
}\right)^{12}-\left(\frac{\mathbf{u}_{ij}+\mathbf{H}_{i j}}{
\left\|\left(\boldsymbol{R} \mathbf{x}_i+\boldsymbol{b}\right)-\left(\boldsymbol{R} \mathbf{x}_j+\boldsymbol{b}\right)\right\|^2
}\right)^6\right] \\
&=\sum_{i, j} \left[\left(\frac{\mathbf{u}_{ij}+\mathbf{H}_{i j}}{
\left(\mathbf{x}_i-\mathbf{x}_j\right)^{\top} \boldsymbol{R}^{\top} \boldsymbol{R}\left(\mathbf{x}_i-\mathbf{x}_j\right)
}\right)^{12}-\left(\frac{\mathbf{u}_{ij}+\mathbf{H}_{i j}}{
\left(\mathbf{x}_i-\mathbf{x}_j\right)^{\top} \boldsymbol{R}^{\top} \boldsymbol{R}\left(\mathbf{x}_i-\mathbf{x}_j\right)
}\right)^6\right] \\
&=\sum_{i, j} \left[\left(\frac{\mathbf{u}_{ij}+\mathbf{H}_{i j}}{
\left(\mathbf{x}_i-\mathbf{x}_j\right)^{\top} \mathbf{I}\left(\mathbf{x}_i-\mathbf{x}_j\right)
}\right)^{12}-\left(\frac{\mathbf{u}_{ij}+\mathbf{H}_{i j}}{
\left(\mathbf{x}_i-\mathbf{x}_j\right)^{\top} \mathbf{I}\left(\mathbf{x}_i-\mathbf{x}_j\right)
}\right)^6\right] \\
&=\sum_{i, j} \left[\left(\frac{\mathbf{u}_{ij}+\mathbf{H}_{i j}}{
\left\|\mathbf{x}_i-\mathbf{x}_j\right\|^2
}\right)^{12}-\left(\frac{\mathbf{u}_{ij}+\mathbf{H}_{i j}}{\left\|\mathbf{x}_i-\mathbf{x}_j\right\|^2}\right)^6\right] \nonumber
\end{aligned}
$}
\end{equation} 
\noindent\makebox[\linewidth]{\hfill$\square$}

\begin{table*}
    \centering
    \begin{tabular}{l|c|c|c|c|c|c|c|c|}
    \hline
    \toprule 
     {Method}& \multicolumn{4}{c|}{CASF-2016 set} & \multicolumn{4}{c|}{CSAR-HiQ set}       \\ 
                           & RMSE (↓)             & MAE (↓)              & SD (↓)              & R (↑)          & RMSE (↓)              & MAE (↓)          & SD (↓)              & R (↑)                       \\ \toprule
    LR \cite{ballester2010machine} & 1.675 (0.000)&	1.358 (0.000)&	1.612 (0.000)&	0.671 (0.000)&	2.071 (0.000)	&1.622 (0.000)	&1.973 (0.000)&	0.652 (0.000)  \\   
    
    SVR \cite{ballester2010machine}   &1.555 (0.000)&	1.264 (0.000)&	1.493 (0.000)&	0.727 (0.000)&	1.995 (0.000)	&1.553 (0.000)&	1.911 (0.000)&	0.679 (0.000)   \\
    
    RF-Score \cite{ballester2010machine}  & 1.446 (0.008)&	1.161 (0.007)&	1.335 (0.010)	&0.789(0.003)&	1.947 (0.012)&	1.466 (0.009)&	1.796 (0.020)&	0.723 (0.007)    \\   
    
    Pafnucy  \cite{stepniewska2018development} &1.585 (0.013) &	1.284 (0.021) &	1.563 (0.022) &	0.695 (0.011) &	1.939 (0.103) &	1.562 (0.094) &	1.885 (0.071) &	0.686 (0.027)    \\
    
    OnionNet \cite{zheng2019onionnet} & 1.407 (0.034) &	1.078 (0.028) &	1.391 (0.038) &	0.768 (0.014)	 &1.927 (0.071)	 &1.471 (0.031) &	1.877 (0.097)	 &0.690 (0.040)    \\   
    
    % GraphDTA \cite{nguyen2021graphdta} & 1.562 (0.022) &	1.191 (0.016) &	1.558 (0.018) &	0.697 (0.008) &	1.980 (0.055) &	1.493 (0.046) &	1.969 (0.057) &	0.653 (0.026)      \\
    
    SGCN \cite{danel2020spatial} & 1.583 (0.033)&	1.250 (0.036)&	1.582 (0.320)&	0.686 (0.015)&	1.902 (0.063)&	1.472 (0.067)&	1.891 (0.077)	&0.686 (0.030) \\
    
    GraphTrans \cite{wu2021representing}  &1.539 (0.044) &	1.182 (0.046) &	1.521 (0.042)	 &0.714 (0.019) &	1.950 (0.072) &	1.508 (0.069) &	1.886 (0.083) &	0.687 (0.033)    \\
    
    NL-GCN \cite{liu2021non} &1.516 (0.019)&	1.198 (0.013)&	1.511 (0.024)&	0.720 (0.010)&	1.840 (0.024)&	1.393 (0.016)&	1.817 (0.028)&	0.716 (0.011) \\

    GNN-DTI \cite{lim2019predicting} & 1.492 (0.025)	 & 1.192 (0.032) & 	1.471 (0.051) & 	0.736 (0.021)	 & 1.972 (0.061) & 	1.547 (0.058) & 	1.834 (0.090) & 	0.709 (0.035) \\

    PIGNet \cite{moon2022pignet} & 1.428 (0.016) &	1.133 (0.009) &	1.425 (0.014) &	0.761 (0.006) &	1.532 (0.026) &	1.198 (0.054) &	1.512 (0.035) &	0.781 (0.013) \\

    MAT \cite{maziarka2020molecule} & 1.457 (0.037) &	1.154 (0.037) &	1.445 (0.033) &	0.747 (0.013)	 &1.879 (0.065) &	1.435 (0.058) &	1.816 (0.083) &	0.715 (0.030) \\

    CMPNN \cite{song2020communicative} & 1.408 (0.028)	 &1.117 (0.031) &	1.399 (0.025) &	0.765 (0.009)	 &1.839 (0.096)	 &1.411 (0.064) &	1.767 (0.103) &	0.730 (0.052) \\

    SIGN \cite{li2021structure} & 1.316 (0.031) & 1.027 (0.025) &1.312 (0.035)  &0.797 (0.012) & 1.735 (0.031) & 1.327 (0.040)  &1.709 (0.044)  &0.754 (0.014) \\

    GIANT \cite{li2023giant} &1.269 (0.020)& 0.999 (0.018)& 1.265 (0.024)& 0.814 (0.008)& 1.666 (0.024) &1.242 (0.030)& 1.633 (0.034) &0.779 (0.011)\\  \hline

    \textbf{SPIN} (Ours)               
    & \textbf{1.258} (0.013)  & \textbf{0.996} (0.021)  & \textbf{1.229} (0.011)    & \textbf{0.826} (0.007)         
    & \textbf{1.288} (0.027)    & \textbf{0.999} (0.034)   & \textbf{1.270} (0.022)   & \textbf{0.800} (0.017) \\ \bottomrule \hline
    \end{tabular}
    \caption{Performance comparision with baselines on the CASF-2016 set and CSAR-HiQ set.}
    \label{table1}
\end{table*}
The optimization strategy of \textbf{SPIN} is derived in two directions from the sum of the previously calculated protein-ligand pairwise interactions (equal to the binding energy) as follows.     
\begin{equation}
\begin{aligned} 
&\hat{y}=\sigma \cdot \mathrm{E}^{\mathtt{VDW}} 
\\& L_\mathtt{d} = \sum_{N}\left(y - \hat{y}\right)^2
\end{aligned}
\end{equation} 
\begin{equation}
\begin{aligned} 
& L_\mathtt{p}= \sum_{N}\sum_{i, j} \left[\frac{\partial\,c_{i j}\left[\left(\frac{\mathtt{r}_i+M_{i j}}{\left\|\mathbf{x}_i-\mathbf{x}_j\right\|}\right)^{12}-2\left(\frac{M_{i j}}{\left\|\mathbf{x}_i-\mathbf{x}_j\right\|}\right)^6\right]}{\partial \, \left\|\mathbf{x}_i-\mathbf{x}_j\right\|}\right]^2
\end{aligned}
\end{equation} 
The first direction is to minimize the mean squared error between the predicted binding energy-derived binding affinity and the experimental binding affinity as shown in Figure.$\text{E}_1$($\sigma$ is a trainable parameter). The second direction involves enforcing a term that satisfies the laws of physics between the protein and ligand, specifically by ensuring the derivative of the binding free energy with respect to the distance between ligand and protein atoms is zero as shown in Figure.$\text{E}_2$. This approach is designed to fulfill the prior knowledge that the modeled complex, as an experimentally elucidated structure, is located at the minimum point of the binding free energy at that reaction coordinate. Ultimately, the total objective function $L$ is defined as $L_\mathtt{d} + L_\mathtt{p}$.

\section{Experiments}
\subsection{Experimental Setup}
% For the binding affinity prediction task, we have set the widely used PDBbind v2020 \cite{liu2017forging} dataset as our training dataset. This dataset provides the 3D crystal structures of protein-ligand complexes with experimentally determined binding affinities. The 3D crystal structures are represented by the coordinates of the atoms that make up the protein and ligand, comprising a total of 13,283 samples. To evaluate the performance of our model and the comparison models, we utilize the CASF-2016 \cite{su2018comparative} and CSAR-HiQ \cite{smith2011csar} benchmark sets. Like PDBbind, these datasets also provide the 3D crystal structures of protein-ligand complexes and their binding affinities. In line with previous studies, CASF-2016 and CSAR-HiQ sets consist of 285 and 343 samples, respectively, and we have avoided duplication in the training dataset by removing these samples from it. During training, we defined the most recently elucidated samples as the validation set, as has been done in recent studies \cite{stark2022equibind}, to verify generalization during the training process.
\noindent{\textbf{Dataset}} For BA prediction, the PDBbind v2020 \cite{liu2017forging} dataset was used as the training dataset and it provided the 3D crystal structures of protein–ligand complexes with experimentally determined BAs. The 3D crystal structures were represented by the coordinates of the atoms constituting the protein and ligand, comprising 19,443 samples. To evaluate the performances of the proposed and comparison models, we used the CASF-2016 \cite{su2018comparative} and CSAR-HiQ \cite{smith2011csar} benchmark sets. Similar to PDBbind, these datasets provide the 3D crystal structures of protein–ligand complexes and their BAs. In accordance with previous studies, the CASF-2016 and CSAR-HiQ sets comprised 285 and 343 samples, respectively, and duplication in the training dataset was avoided by removing these samples. During training, the most recent samples were defined as the validation set to verify the generalizability, as has been done in recent studies \cite{stark2022equibind}.

\noindent{\textbf{Implement detail}} The key/value/query embedding in the SE(3)-invariant graph transformer, a component of SPIN, is obtained through a 2-layer MLP with layer normalization and ReLU activation. The transformer has 16 layers, and the hidden dimension and the number of heads are each 128 and 9, respectively. Additionally, the swish \cite{ramachandran2017searching} function was used as the activation function for each layer.
We have implemented an exponential decay of the learning rate, using a factor of 0.6 and setting a minimum of 1e-6 for the learning rate with Adam. If the validation loss does not improve over 20 consecutive evaluations, the learning rate is reduced. Our model training was conducted on a single NVIDIA GeForce GTX 3090 GPU.

\subsection{Performance Evaluation}
% We first compare the proposed \textbf{SPIN} with existing baseline models on two benchmark datasets. 
First, the proposed \textbf{SPIN} was compared with the existing baseline models on two benchmark datasets. 
As shown in Table 1, it achieves the best performance across all four metrics for the two benchmark datasets.  
% It can be observed that CNN-based methods such as Pafnucy\cite{stepniewska2018development} and OnionNet\cite{zheng2019onionnet} lack generalization ability on benchmark datasets because they do not satisfy the invariance to rotations and translations of the complex. 
CNN-based methods, such as Pafnucy \cite{stepniewska2018development}, lack generalizability on benchmark datasets because they are not invariant to rotations and translations of the complex. 
% One of the GNN-based methods, GraphDTA\cite{nguyen2021graphdta}, which does not utilize the 3D structure of the complex, fails to model structural information, resulting in inferior performance compared to CNN-based models.
SGCN\cite{danel2020spatial}, a GNN-based method utilizes spatial structures but directly inputs coordinates for predictions, encounters similar issues as CNN-based models and is sensitive to rotations and translations, thus exhibiting comparable performance levels. The Graphtrans\cite{wu2021representing}, NL-GCN\cite{liu2021non}, PIGNet\cite{moon2022pignet}, and CMPNN\cite{song2020communicative} models are not ideal for BA prediction because they rely solely on connectivity information without modeling the geometric characteristics of the 3D space. The SIGN\cite{li2021structure} and GIANT\cite{li2023giant} models incorporate various modules to better capture the spatial information of proteins and ligands by considering distance and angle information. However, since they follow a method based on predefined rules obtained from training data, they are not flexible enough to handle various complex structures that are not present in the training data. 
Especially, the result that the proposed model, \textbf{SPIN}, achieved a 30$\%$ improvement over the best baseline models on the CSAR set, which is comparatively difficult to generalize across all comparison models, suggests that injecting inductive biases from various perspectives into the prediction model plays a crucial role in generalizing to unseen datasets.
An interesting observation is that another model, PIGNet\cite{moon2022pignet}, which defines physicochemical laws as the 
inductive bias of the prediction model, also achieved relatively high performance on the challenging CSAR set compared to other comparison models.
This reaffirms the importance of defining physicochemical information as the inductive bias of the prediction model from the perspective of applications that require predicting binding affinity for various complexes. 
However, because this model does not clearly model the spatial information of complex structures, it still shows inferior performance in the CASF-2016, indicating it is not ideal as a model for predicting binding affinity.

PIGNet, which defines physicochemical laws as the inductive bias of the prediction model, significantly outperformed the other models on the challenging CSAR set. This validates the importance of defining physicochemical information as the inductive bias of the prediction model based on BA prediction for various complexes. PIGNet did not represent the spatial information of complex structures; however, it exhibited inferior performance on CASF-2016, indicating its unsuitability for BA prediction.

\begin{figure}
	\centering
    \includegraphics[width=3.37in]{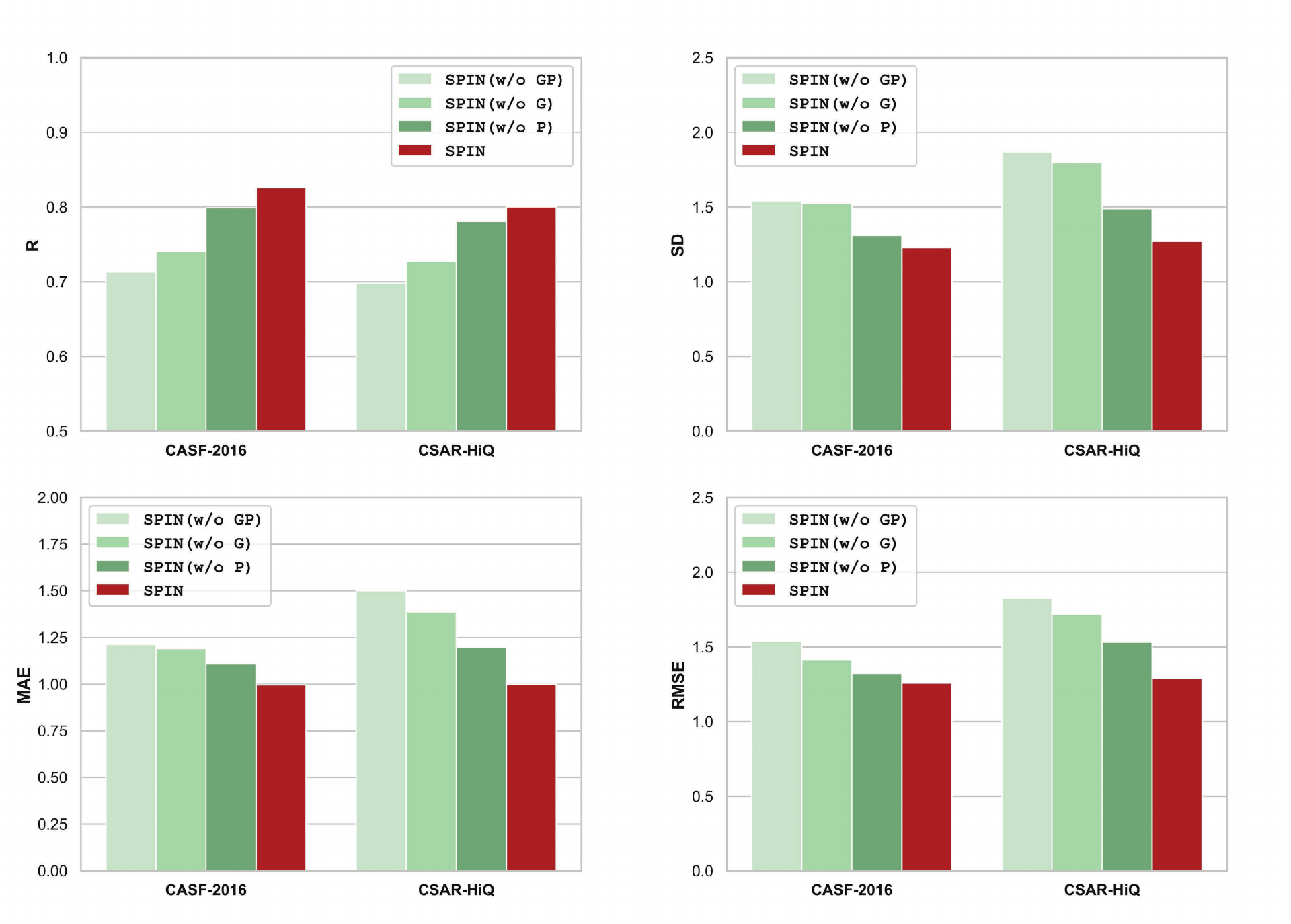}
	\caption{Ablation studies on CASF-2016, CSAR HiQ set. Performance for four evaluation metrics are presented for four cases of the SPIN model: the complete model with all inductive biases injected (\texttt{SPIN}), the case without geometric inductive bias (\texttt{SPIN[w/o G]}), the case without physicochemical inductive bias (\texttt{SPIN[w/o P]}), and the case with both biases removed (\texttt{SPIN[w/o GP])}).}
	\label{fig3}
\end{figure}
\subsection{Ablation Study}
In this section, we conduct an ablation study on the \textbf{SPIN} model to verify the importance of each component. \texttt{SPIN(w/o G)} is a model from which the geometric inductive bias that satisfies SE(3)-invariance has been removed, \texttt{SPIN(w/o P)} is a model from which the physicochemical inductive bias concerning the Binding free energy minimum has been removed, and finally, \texttt{SPIN(w/o GP)} is a model with both types of inductive bias removed. We measured the performance of these four variant models, including the complete \texttt{SPIN} model, on two benchmark datasets. The results, as shown in Figure 3, confirm that both geometric and physicochemical inductive biases play essential roles in predicting binding affinity. Specifically, when comparing the performance of \texttt{SPIN(w/o G)} and \texttt{SPIN(w/o P)}, it is revealed that the geometric inductive bias, namely the condition that the binding affinity remains invariant to the rotation and translation of the protein-ligand complex, is the most critical component for prediction performance. It is an interesting observation that the physicochemical inductive bias shows better synergy when conditions for reasonably modeling the complex's geometric information are met. This outcome, considering that the binding free energy calculations are based on the complex’s position in three-dimensional space, provides significant insights into the interplay between the two types of inductive biases.

\subsection{Virtual Screening}
% While predicting the binding affinity of a protein-ligand complex is critically important, in drug development it is also essential to correctly rank multiple ligands based on their binding strength to the target protein. This allows for the proper listing of the most promising drug candidates, thereby making the drug development process more efficient.
Along with predicting the BA of a protein–ligand complex, correctly ranking multiple ligands based on their binding strength to the target protein is crucial in drug development. This facilitates the proper listing of the most promising drug candidates, thereby making the drug development process more efficient. 
For this purpose, we adopt the methodology of previous studies \cite{li2014comparative, su2018comparative} about CASF-2016 benchmark set to validate the practicality of \textbf{SPIN}. 
Specifically, we measure the ranking power, which is the ability of a scoring function to accurately rank known ligands of a specific target protein based on binding affinity when the precise binding pose is given. This approach allows us to rigorously assess the efficacy of \textbf{SPIN} in predicting binding affinities.
The CASF2016 set comprises 57 protein clusters, each cluster containing five complexes that are bound to the same protein but exhibit markedly different binding affinities. Each cluster has pre-defined rankings based on the binding affinities for different ligands. 
In each cluster, rankings are established based on the binding affinities for different ligands, and we measure how accurately the predictive model inferences these rankings using the Spearman’s rank correlation coefficient. 
The average of these coefficients across all clusters is defined as the ranking power for this experiment.
We selected several scoring functions that are utilized in actual docking protocols as comparative models. Additionally, the Random Forest model, already trained on the CASF-2016 dataset, was excluded from this comparison \cite{su2018comparative}.
As demonstrated in Figure 4, \textbf{SPIN} exhibits superior ranking power compared to the scoring functions of other docking programs. This indicates that our proposed model can effectively prioritize candidate compounds in real-world drug development scenarios.
\begin{figure}
	\centering
    \includegraphics[width=3.37in]{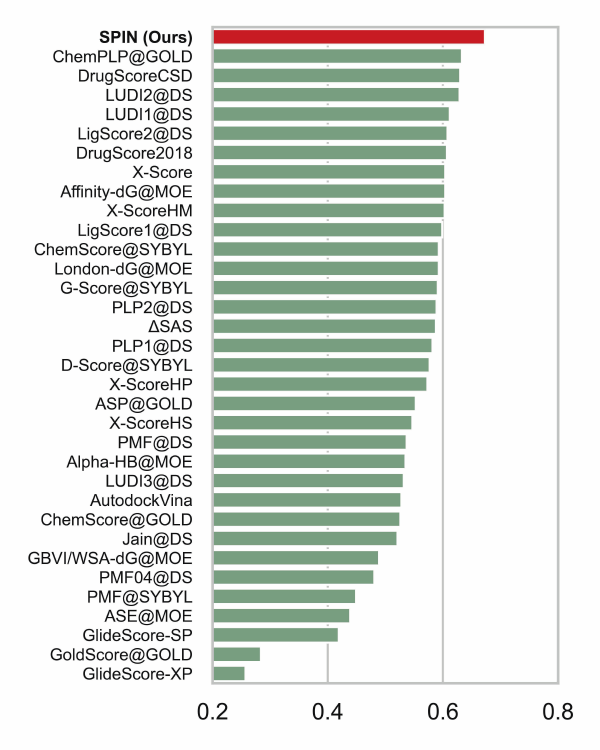}
	\caption{Average Spearman correlation coefficient obtain on 57 target proteins by each scoring function included proposed \textbf{SPIN} in the ranking power test.}
	\label{fig3}
\end{figure}
\subsection{Interpretability}
To determine if the predictions output by a predictive model can be actively utilized in the drug development process with sufficient reliability, it is crucial to analyze the interpretability of the predictive model. 
% No matter how excellent the predictive performance, in actual drug development, if the basis for the predicted binding affinity is not clear, it can be challenging to trust the results. This lack of clarity can become an obstacle to efficient development in subsequent processes, such as lead optimization \cite{jorgensen2009efficient}.
Independent of the predictive performance in actual drug development, result validation can be challenging if the basis of the predicted BA is ambiguous, hindering the efficient development of subsequent processes, such as lead optimization \cite{jorgensen2009efficient}.
To evaluate this, we utilize the pairwise interaction matrix $\mathbf{H}$ within the \textbf{SPIN} framework, which involves interactions between protein and ligand atoms. The protein-ligand interaction matrix $\mathbf{H}_{ij}$ represents the binding energy between protein atom $i$ and ligand atom $j$, where a lower value indicates stronger interaction between the two substructures. We extract the amino acids of the protein corresponding to the lowest 10$\%$ of the energy values in the interaction matrix output by the trained predictive model.
To substantiate the interpretability of \textbf{SPIN}, we focus our analysis on the 3bu1(PDB ID) protein-ligand complex, as illustrated in Figure 5.(a), by visualizing amino acids that engage in strong interactions. To confirm that these extracted amino acids are indeed engaged in actual intermolecular interactions, we compare our results with the findings from Discovery Studio's intermolecular interaction profiler. The amino acids corresponding to the lowest 10$\%$ of the energy values in \textbf{SPIN}’s interaction matrix were specifically \textit{21.A TYR, 37.A VAL, 51.A TYR, 94.A ASP, 96.A VAL, and 105.A TRP} ( [\textit{residue index}].\textit{A} [\textit{amino acid}] ). These residues are hypothesized to be pivotal in the protein-ligand interactions during the predictive modeling performed by \textbf{SPIN}.
\begin{figure}
	\centering
    \includegraphics[width=3.37in]{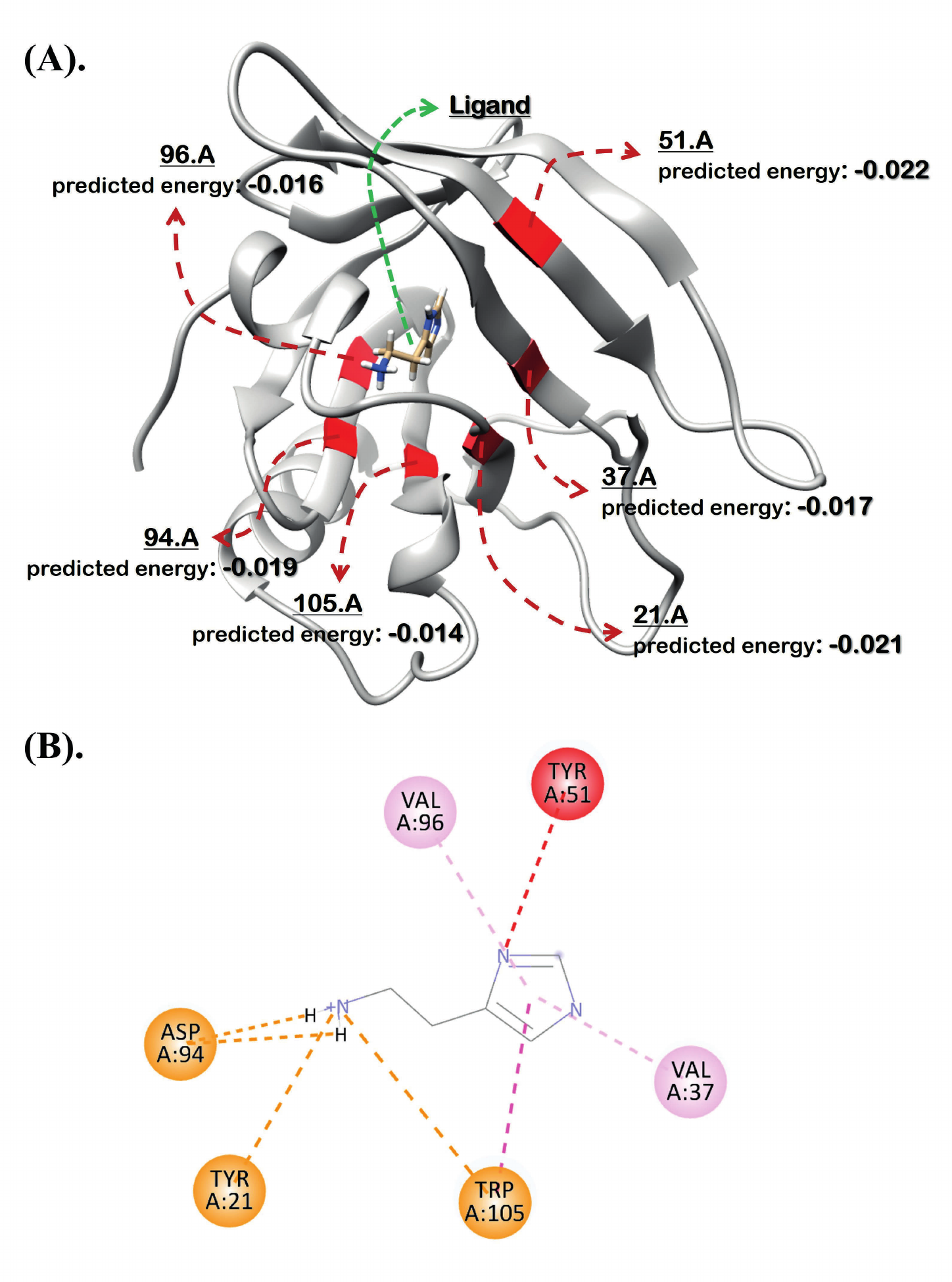}
	\caption{(A). Visualization of the 3bu1 PDB sample: Protein amino acids corresponding to the lowest 10$\%$ of the predicted protein-ligand interaction energy are highlighted in red. Additionally, the specific energy values are annotated alongside these amino acids. (B). Results from the Discovery Studio interaction profiler for the 3bu1 PDB sample show that a total of six amino acids are involved in critical interactions.}
	\label{fig4}
\end{figure}
Subsequent analysis confirmed that these amino acids matched precisely with those identified in Discovery Studio’s profiling results, as depicted in Figure 5.(b). This concordance strongly indicates that \textbf{SPIN} not only accurately predicts binding affinity but also reliably identifies the biologically relevant interactions that underpin these predictions. This capability to discern and rationalize complex interactions in biological systems underscores the robust interpretability of \textbf{SPIN}, affirming its utility in predictive modeling within the biochemical research field. This validation lends significant credence to \textbf{SPIN}’s application in computational drug discovery, demonstrating its potential to contribute effectively in the field by providing insights that are both predictive and interpretable.

\section{Conclusion}
In this work, we propose SPIN, a binding affinity prediction model infused with various inductive biases, designed to achieve excellent generalization performance from limited data. The model incorporates a geometric inductive bias that assumes binding affinity remains constant regardless of rotations and translations in three-dimensional space, and a physicochemical inductive bias that posits binding occurs at minimal binding free energy. Through rigorous validation across multiple benchmark sets, the superiority of our proposed model is confirmed. Additionally, the model's practicality in virtual screening during actual drug development processes is demonstrated. Finally, by visualizing the interpretability of the prediction model, we ensure the reliability of the values it predicts. The proposed model shows potential in the screening process for selecting molecules that strongly bind to a target protein among various molecular structures. Furthermore, using this prediction model to generate molecules with desired properties in the field of generative models presents an interesting research direction.

\begin{ack}
This research was supported by the National Research Foundation (NRF) funded by the Korean government (MSIT) (No. RS-2023-00229822). The funders did not play any role in the design of the study, data collection, analysis, or preparation of the manuscript.
\end{ack}

%%% Use this environment to include acknowledgements (optional).
%%% This will be omitted in doubleblind mode.

% \begin{ack}
% By using the \texttt{ack} environment to insert your (optional) 
% acknowledgements, you can ensure that the text is suppressed whenever 
% you use the \texttt{doubleblind} option. In the final version, 
% acknowledgements may be included on the extra page intended for references.
% \end{ack}

%%%%%%%%%%%%%%%%%%%%%%%%%%%%%%%%%%%%%%%%%%%%%%%%%%%%%%%%%%%%%%%%%%%%%%%%

%%% Use this command to include your bibliography file.

\bibliography{ecai-sample-and-instructions}
\end{document}